%% file: main.tex
\documentclass[sigconf]{acmart}

\def\BibTeX{{\rm B\kern-.05em{\sc i\kern-.025em b}\kern-.08emT\kern-.1667em\lower.7ex\hbox{E}\kern-.125emX}}

\usepackage{amsmath}
\usepackage{algorithm}
\usepackage{algorithmic}

\copyrightyear{2020}
\acmYear{2020}
\setcopyright{rightsretained}
\acmConference[KDD '20] {26th ACM SIGKDD Conference on Knowledge Discovery and Data Mining}{August 23--27, 2020}{Virtual Event, USA}
\acmBooktitle{26th ACM SIGKDD Conference on Knowledge Discovery and Data Mining (KDD '20), August 23--27, 2020, Virtual Event, USA}
\acmPrice{}
\acmDOI{10.1145/3394486.3403372}
\acmISBN{978-1-4503-7998-4/20/08}
\begin{document}
\fancyhead{}

%
\title{Shop The Look: Building a Large Scale Visual Shopping System at Pinterest}

%

\author{Raymond Shiau, Hao-Yu Wu, Eric Kim, Yue Li Du, Anqi Guo, Zhiyuan Zhang, Eileen Li, \and Kunlong Gu, Charles Rosenberg, Andrew Zhai}
\affiliation{Pinterest, Inc.\\San Francisco, CA}
\email{{rshiau,rexwu,erickim,shirleydu,aguo,zhiyuan,eileenli,kgu,crosenberg,andrew}@pinterest.com}

%

\renewcommand{\shortauthors}{Shiau et al.}

%

\include{main_abstract}

%
%


%

%

%
\maketitle


\input{main_introduction.tex}

\input{main_system.tex}
\input{main_evaluation.tex}
\input{main_decomposition.tex}

\input{main_embeddings.tex}
\input{main_label_collection.tex}
\input{main_lessons.tex}
\input{main_conclusion.tex}
\input{main_acknowledgements.tex}

%

\bibliographystyle{ACM-Reference-Format}
\bibliography{sample-base}

\clearpage

%
\appendix

\input{appendix.tex}

\end{document}

%% file: main_abstract.tex
\begin{abstract}
As online content becomes ever more visual, the demand for searching by visual queries grows correspondingly stronger. Shop The Look is an online shopping discovery service at Pinterest, leveraging visual search to enable users to find and buy products within an image. In this work, we provide a holistic view of how we built Shop The Look, a shopping oriented visual search system, along with lessons learned from addressing shopping needs. We discuss topics including core technology across object detection and visual embeddings, serving infrastructure for realtime inference, and data labeling methodology for training/evaluation data collection and human evaluation. The user-facing impacts of our system design choices are measured through offline evaluations, human relevance judgements, and online A/B experiments. The collective improvements amount to cumulative relative gains of over 160\% in end-to-end human relevance judgements and over 80\% in engagement. Shop The Look is deployed in production at Pinterest.
\end{abstract}

%% file: main_introduction.tex
\section{Introduction}

\begin{figure}[t]
\begin{center}
\includegraphics[width=1.0\linewidth]{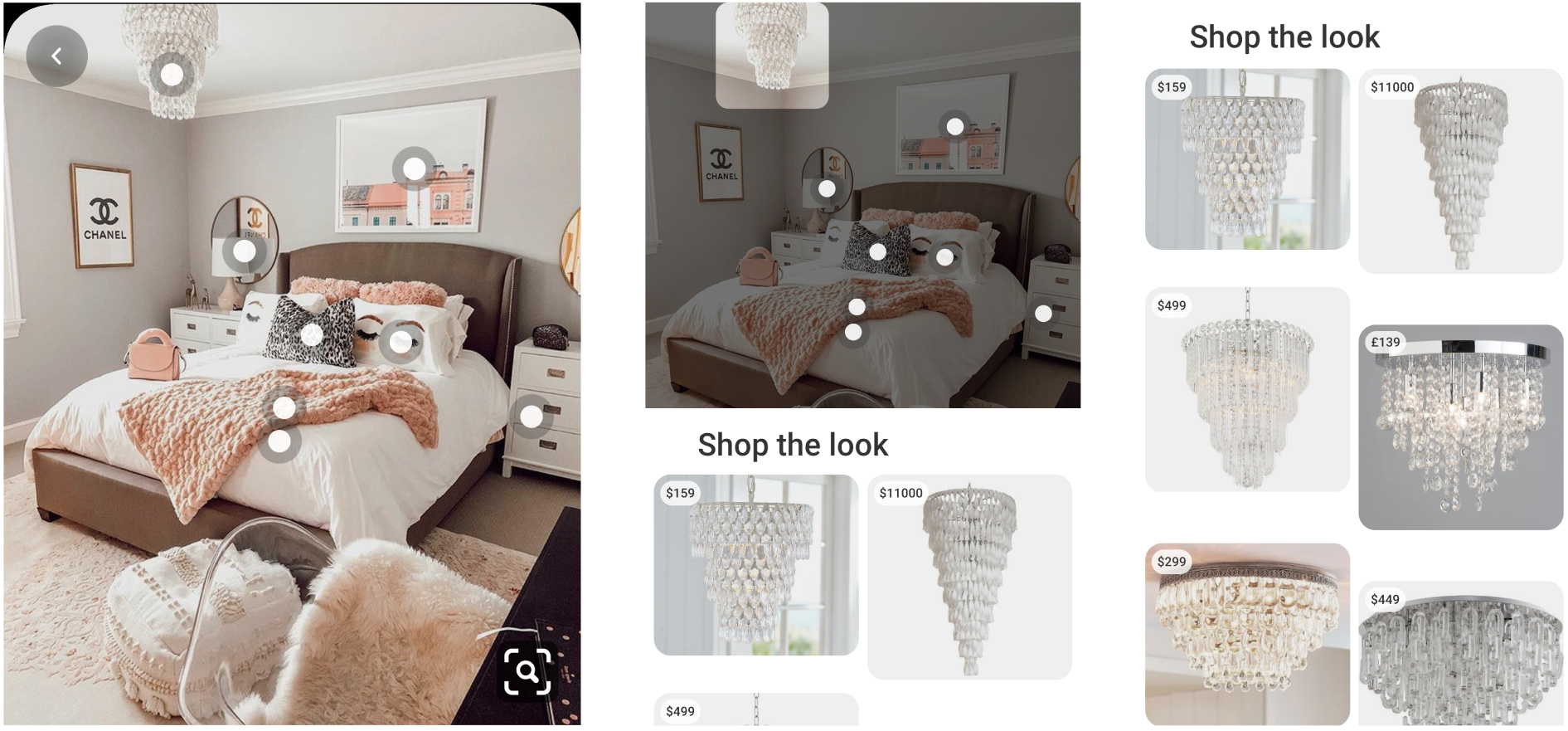}
\end{center}
\caption{Example of Shop The Look. Detected objects in a scene are displayed as dots (left). Tapping on a dot expands it into a bounding box and displays Shop The Look product results (middle). Swiping up shows the full Shop The Look product feed (right).}
\label{fig:ui}
\vspace{-2mm}
\end{figure}

\begin{figure*}[h!]
\includegraphics[width=1.0\linewidth]{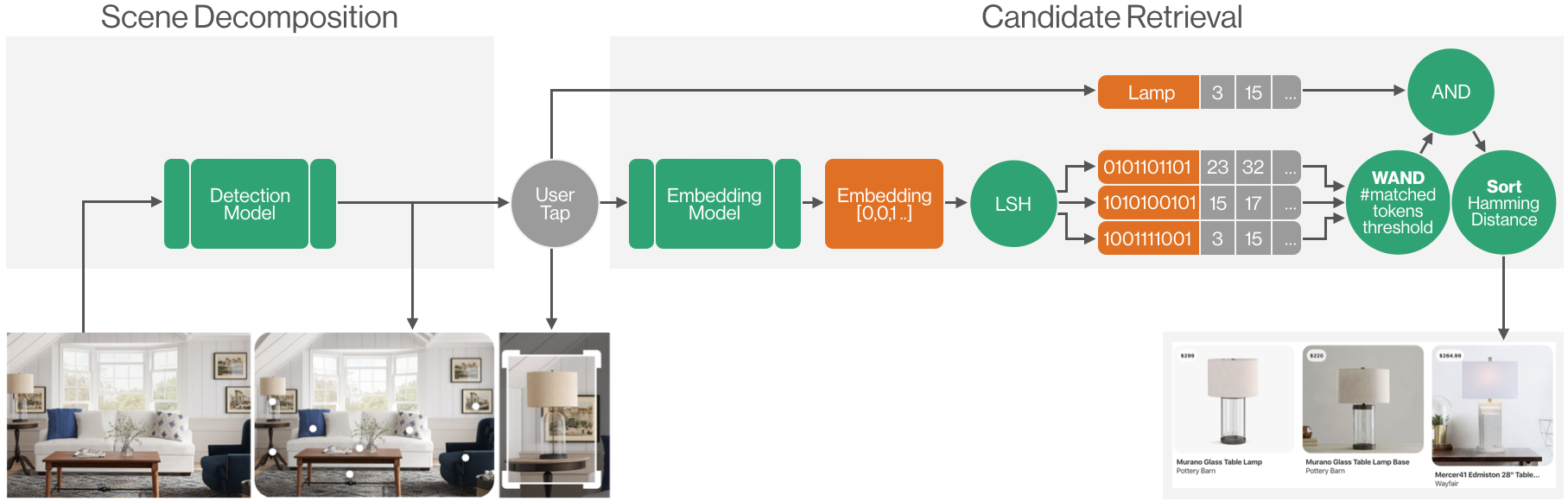}
\vspace{-5mm}
\caption{Overview of the Shop The Look system.}
\label{fig:overview}
\vspace{-2mm}
\end{figure*}

With the proliferation of visual information online in the form of photos and videos, visual search has become an increasingly popular means to search for information that is hard to articulate through text. 
A more recent trend is enabling product search through visual queries (\textit{visual shopping}), particularly in categories such as fashion and home decor where visual attributes are critical in describing the product. Visual shopping systems have been built by Amazon \cite{amazon-stylesnap}, Alibaba \cite{alibaba}, eBay \cite{ebay}, Google \cite{google-lens}, and Microsoft \cite{microsoft-bing}.
Among over 350M Pinterest monthly users, many have also expressed a desire for visual shopping. To address this need, we built Shop The Look, a visual shopping system that detects objects within billions of inspirational scenes, and finds matching products from our corpus of hundreds of millions of products that are visually similar to those objects (Figure~\ref{fig:ui}).


Our contribution is providing a view into the evolution of a visual shopping system from an initial prototype to a production quality system. We first present the evaluation methodology for Shop The Look (Section~\ref{sec:evaluation}), which serves as the basis of improvements for all the components in the system. We show the evolution of the building blocks of Shop The Look (Figure \ref{fig:overview}) -- \textbf{scene decomposition (detection)} (Section~\ref{sec:decomposition}), \textbf{candidate retrieval (embedding)} (Section~\ref{sec:embeddings}) -- through system infrastructure, model, and most importantly, dataset improvements. The importance of \textbf{label collection} (Section~\ref{sec:label_collection}) motivates us to invest in our labeling platform for both training/evaluation data collection and human relevance evaluation. 
Many works on visual search systems document a single snapshot of the system in time. We, however, describe the evolution of our system across dataset, modeling, and infrastructure with justifications for each step. We examine the impacts of these components, and show how we improve them over time by addressing specific needs for visual shopping. We demonstrate user-facing impact through offline evaluations, human relevance evaluations, and online A/B experiments (Section~\ref{subsec:detection_exps}, \ref{subsec:embedding_exps}), and we reflect on practical lessons learned (Section~\ref{sec:lessons}). By describing in detail the journey, we hope our contributions will give insights into how to improve an industry-scale visual shopping application for other visual shopping practitioners.



\input{introduction_related_works.tex}

%% file: introduction_related_works.tex
\section{Related Works}
\subsection{Visual search systems}

Due to the rapid advancements in visual understanding through deep learning, many companies have invested in building visual search systems~\cite{ebay, alibaba, amazon-stylesnap, google-lens, jd, microsoft-bing, unified-embedding}. Shopping is a core focus in visual search across production systems~\cite{ebay, alibaba, jd, amazon-stylesnap} and academic research~\cite{bell15productnet, fashion_rec}. Many describe visual search systems at a single point in time. Few detail the evolution of the system and how to address heterogeneous requirements for a specific use case such as shopping.



\subsection{Representation Learning}
Image representation, or so-called image embedding, is commonly trained with deep metric learning algorithms. Standard deep metric learning aims to pull together positive examples while pushing apart negative examples in the learned embedding space. The challenges of such algorithms lie in the need for large batch size and careful negative sampling \cite{sampling-matters}. An alternative approach is based on classification loss, which alleviates those challenges, and has recently achieved state-of-the-art results \cite{classification-embedding}. It also opens up the possibility of incorporating multiple objectives in training a single embedding \cite{unified-embedding}.

\subsection{Object Detection}
Object detection detects where objects of interest are, and their corresponding category labels. The state-of-the-art object detectors are two-stage with architecture resembling FasterRCNN \cite{DBLP:journals/corr/RenHG015}. The development of Feature Pyramid Network \cite{DBLP:journals/corr/LinDGHHB16} and data augmentation techniques have improved detection performance for objects at various scales.

%% file: main_system.tex

\section{System Overview}

Shop The Look has two main modules: scene decomposition (detection) and candidate retrieval (embedding). Figure~\ref{fig:overview} depicts how the two modules fit together to power Shop The Look.

\textbf{Scene decomposition (detection).} For scene images, we use an object detector to identify the prominent objects in the image. Detected objects are displayed in the UI as white dots that users can tap on, indicating their intent to "Shop The Look" or \textit{shop} a particular object in the scene. The detector outputs a bounding box and category label for each detected object. Recall of object detection impacts the volume of total objects that become \textit{shoppable} (i.e. eligible for Shop The Look). Precision is important to identify accurate bounding boxes and correct category labels used for filtering Shop The Look product results.

\textbf{Candidate retrieval (embedding).} For the scene image we decomposed into \textit{shoppable} objects, the user can tap on an object to get visually similar product recommendations. We extract the visual embedding of the object crop and fetch its nearest neighbor results, filtered by predicted category label. To improve visual embeddings for the shopping use case, we build upon the multi-task Unified Embedding~\cite{unified-embedding} at Pinterest. We add new tasks to improve recommendation relevance and address specific failure modes. This directly influences the visual similarity (relevance) of retrieved candidates.


During the development of the two modules, we uncovered challenges in \textbf{label collection}, along with its importance for monitoring and providing improvements. How the training data is annotated dictates the behavior of our system, and training data quality directly impacts model performance. Furthermore, we evaluate end-to-end system relevance based on human judgements. Reliable and consistent labeling is the key to accurate measurement of system improvements. We have heavily invested in data collection tools and processes that allow us to quickly and reliably obtain high quality labels.

%% file: main_evaluation.tex
\section{Evaluation Methodology}
\label{sec:evaluation}
\begin{figure}[h!]
\begin{center}
\includegraphics[width=0.7\linewidth]{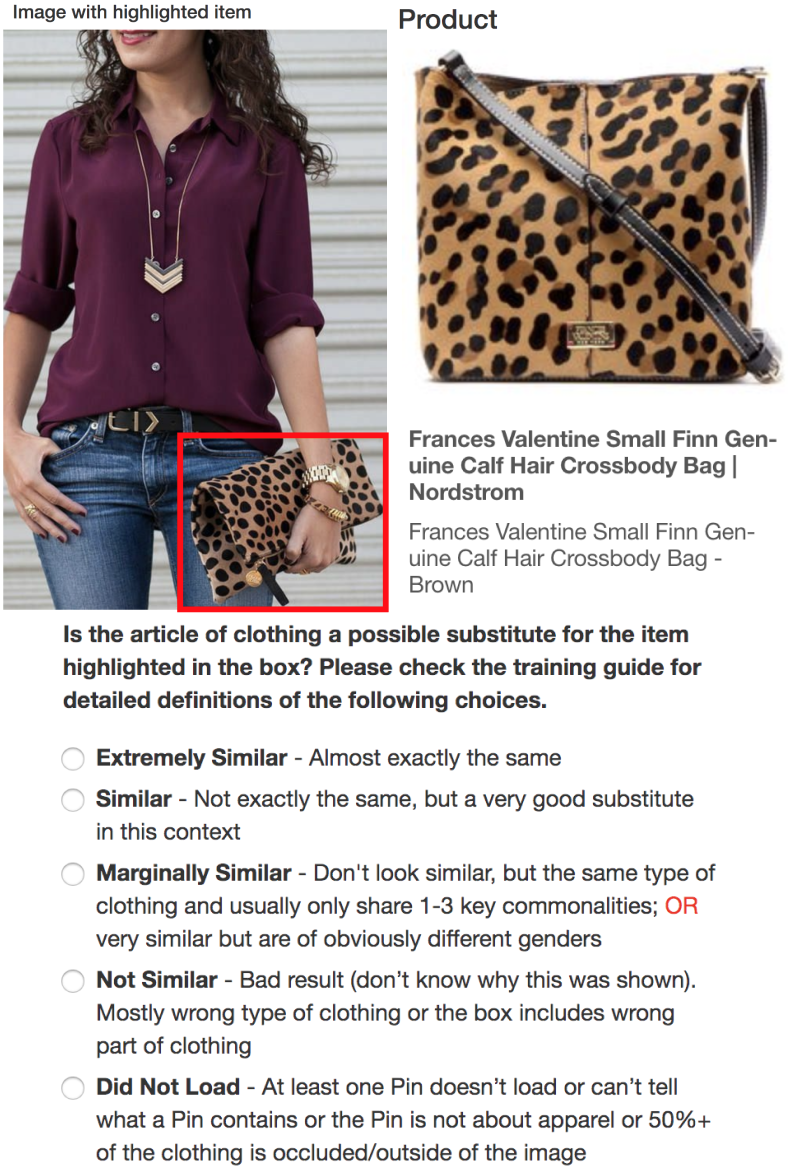}
\end{center}
\caption{Example question given to human raters for the ``End-to-end Relevance'' human evaluation task.}
\label{fig:e2e_fashion}
\vspace{-2mm}
\end{figure}

The goal for Shop The Look is to fulfill the user's shopping intent. We evaluate our system via offline evaluations, human relevance evaluations, and online engagement A/B experiments. The most important metrics for measuring the success of the system are: \emph{scene closeup volume}, \emph{product click-through volume}, and \emph{product click-through rate} from online engagement A/B experiments; and \emph{End-to-end (E2E) Relevance@5} from human relevance evaluations.
\begin{itemize}

\item{\textit{Scene closeup volume.}} This metric measures the number of scene impressions that contain detected objects. It indicates how many times scenes viewed by users contained \textit{shoppable} objects, and is an indicator for Shop The Look coverage.

\item{\textit{Product click-through volume.}} This metric measures the number of clicks on products. It indicates how many times users engaged with product results by clicking on the link to visit the product's website, and is an indicator for total shopping engagement on Shop The Look product results.

\item{\textit{Product click-through rate.}} This metric measures \textit{product click-through volume / product impression volume}. It indicates how often the products recommended by Shop The Look resulted in a click (to linked website), and is an indicator for Shop The Look product result quality.



\item{\textit{End-to-end (E2E) Relevance@5.}} This metric measures the relevance of the top-5 candidates generated by scene decomposition and embedding retrieval.
We first run the scene decomposition module on a set of Shop The Look scene images.
For each query object from scene decomposition, we take the top-5 candidate products output by embedding retrieval and ask human raters to verify if the candidate product is relevant to the query object (Figure~\ref{fig:e2e_fashion}). Relevance ratings can be either "Extremely Similar": the candidate is almost the same as the query, "Similar": the candidate has minor differences to the query such as color or shape, or "Bad" (including "Marginally Similar", "Not Similar", or "Did Not Load" in the figure).
From these answers, we calculate the top-5 candidate precision aggregated over all query objects.

\end{itemize}
Though the four metrics above are the core metrics we aim to improve, our modeling experiments are mostly evaluated offline for practical reasons. For instance, to run full-scale embedding experiments, we must compute embeddings for hundreds of millions of product images and index them in the retrieval system. As another example, it can take weeks to gather significant engagement data for an A/B experiment. For faster experiment iteration, we collect offline evaluation datasets for both detection (Section~\ref{subsubsec:detection_offline}) and embedding (Section~\ref{subsubsec:embedding_offline_eval}) models. We will detail in the later sections how we design our offline evaluations to simulate the metrics described above. In practice, we find there is a positive correlation between gains in offline metrics and relevance/engagement metrics.

%% file: main_decomposition.tex
\section{Scene Decomposition (Detection)}
\label{sec:decomposition}




To recommend shopping products that are highly relevant to a particular image, we first analyze the image and detect the objects that are present, as well as the spatial regions of each detected object.
This is implemented by running an object detection model on the image, which outputs a category label and spatial bounding box for each detected object.
These objects are used as visual queries in Shop The Look.

Our detectors have a two-stage architecture similar to FasterRCNN \cite{DBLP:journals/corr/RenHG015}, consisting of a Region Proposal Network followed by an object classification loss and a bounding box regression loss. The detection training set comprises 251,000 training images with 727,000 bounding boxes in the home decor and fashion domains. We set aside a validation set consisting of 10,000 images with 30,000 bounding boxes for tuning detection thresholds. See Appendix~\ref{subsec:dtn_setup} for further details on our detection training setup.



\subsection{Detection improvements}

The object detector has critical impact on Shop The Look coverage and result quality. 
Thus, we invest in several improvements to the core object detection technology:
\begin{enumerate}
    \item We improve detection performance on single-object queries because they constitute a significant portion of Shop The Look queries overall, through creating a single-product image dataset for training (Section~\ref{subsubsec:single_product}).
    \item We upgrade our detector to more advanced detection model architectures in the literature (Section~\ref{subsubsec:detection_architecture}).
    \item To improve performance for objects of various scales, we enable multi-scale image augmentation (Section~\ref{subsubsec:multiscale}).
\end{enumerate}





\input{detection_single_product.tex}
\subsubsection{\textbf{Detection architecture}}
\label{subsubsec:detection_architecture}
Since the release of the object detector for the general purpose visual search system at Pinterest \cite{pinterest-visual-search}, there have been advances in object detection technology that improve the performance of detection models.
We upgrade our detector from a ResNet101-FasterRCNN \cite{DBLP:journals/corr/HeZRS15} \cite{DBLP:journals/corr/RenHG015} model to a state-of-the-art ResNext101-FasterRCNN \cite{DBLP:journals/corr/XieGDTH16} with Feature Pyramid Networks (FPN) \cite{DBLP:journals/corr/LinDGHHB16} model.

\subsubsection{\textbf{Multi-scale training}}
\label{subsubsec:multiscale}
We enable multi-scale image augmentation during training, by resizing the ground truth image and box coordinates to a randomly-chosen target image size, while maintaining the aspect ratio.
This improves robustness to object scale, which is important in the shopping application as the same product appears at different scales depending on the context of the image.








\input{detection_experiments.tex}

%% file: detection_single_product.tex
\subsubsection{\textbf{Single-product dataset}}
\label{subsubsec:single_product}

\begin{figure}[t]
\begin{center}
\includegraphics[width=0.75\linewidth]{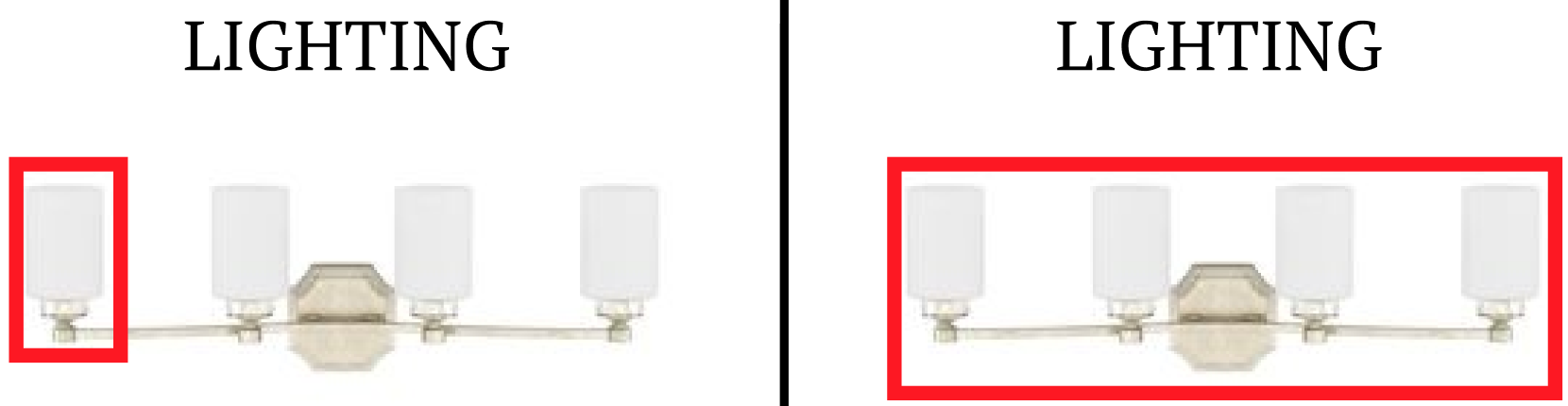}
\end{center}
\vspace{-3mm}
\caption{Visualization of improved detection results on single-product images. We show the detected bounding box and category before (left) and after (right) adding the single-product image dataset into detection training.}
\label{fig:single_product_vis}
\vspace{-2mm}
\end{figure}

Single-object queries such as products on white backgrounds comprise \char`\~39\% of potential \textit{Shop The Look Home} and \textit{Fashion} query images. 
Typically, detection datasets (such as COCO \cite{DBLP:journals/corr/LinMBHPRDZ14}) primarily consist of scenes containing multiple objects, and don't have a representative sample of single-object images.
Because we want the option to trigger Shop The Look on single-object queries (Figure~\ref{fig:single_product_vis}), we create a weakly supervised single-product image dataset with 133,000 images from our product corpus, taking advantage of merchant-provided category labels and bootstrapping bounding boxes using our detection model (See Appendix~\ref{subsec:single_product_generation} for pseudocode). Though the resulting bounding boxes can be somewhat noisy, our approach does not require human curation, which dramatically speeds up the development cycle. 

%% file: detection_experiments.tex
\subsection{Detection experiments}
\label{subsec:detection_exps}
We explore the following detection model variants (Table~\ref{tab:detection_variants}):

\begin{description}
  \item[Baseline.] Faster-RCNN \cite{DBLP:journals/corr/RenHG015} with a ResNet101 backbone.
  \item[+SPI.] Augmented the training set with single-product images (``+Single-Product''), as in Section~\ref{subsubsec:single_product}.
  \item[+FPN.] Upgraded the backbone to ResNeXt101 \cite{resnext}, and added Feature Pyramid Networks \cite{DBLP:journals/corr/LinDGHHB16}, as in Section~\ref{subsubsec:detection_architecture}. Further, due to the category taxonomy growing more comprehensive over time with the needs of Shop The Look, we re-labeled the training set to migrate from coarse-grained to fine-grained Google Product Taxonomy (GPT) categories (``+Fine-grained'').
  \item[+MUL +CLN.] In addition to multi-scale image augmentation during training (Section~\ref{subsubsec:multiscale}), we cleaned up fashion categories in the training set to address label quality issues (``+Fashion Cleanup''). \footnote{Because home decor categories were untouched by the dataset cleanup, and detection evaluations focus on home decor categories, we believe that any gains for the ``+MUL +CLN'' model variant are primarily due to the multi-scale image augmentation.}
\end{description}

For brevity, we only report results on home decor relevance for detection experiments.

\begin{table}
\begin{center}
\begin{tabular}{ |c|c|c|c| } 
 \hline
  Model Name & Architecture & Train Dataset \\
 \hline
 Baseline & ResNet101-FasterRCNN & - \\ 
 +SPI & ResNet101-FasterRCNN & +Single-Product \\ 
 +FPN & ResNeXt101-FasterRCNN-FPN & +Fine-grained \\ 
 +MUL +CLN & ResNeXt101-FasterRCNN-FPN & +Fashion Cleanup \\
 \hline
\end{tabular}
\caption{Summary of detection model variants.}
\label{tab:detection_variants} 
\end{center}
\vspace{-8mm}
\end{table}

\begin{table}
\begin{center}
\begin{tabular}{l r r r r r r}
\hline
Model & \multicolumn{3}{c}{Scene Images} &  \multicolumn{3}{c}{Single-Product Images} \\
 & mAP & P & R & mAP & P & R \\
\hline\hline
Baseline & 0.285 & 0.554 & 0.313 & 0.102 & 0.199 & 0.131 \\
+SPI & 0.276 & 0.523 & 0.285 & 0.154 & 0.188 & 0.139 \\
+FPN & 0.373 & 0.528 & 0.407 & 0.131 & 0.205 & 0.167 \\
+MUL +CLN & 0.444 & 0.563 & 0.527 & 0.265 & 0.372 & 0.392 \\
\hline
\end{tabular}
\end{center}
\caption{Detection model experiments on offline evaluations. mAP: Mean average precision, P: Precision, R: Recall.}
\label{tab:detection_offline}
\vspace{-5mm}
\end{table}

\subsubsection{\textbf{Offline evaluation}}
\label{subsubsec:detection_offline}
We collect two different test sets for the offline detection evaluation: ``Scene Images'', which are images of home decor scenes like Figure~\ref{fig:ui} (1,200 images), and ``Single-Product Images'', which are images of single products like Figure~\ref{fig:single_product_vis} (1,400 images). We collect bounding boxes and category labels for visual objects in each image. Three metrics are reported in Table ~\ref{tab:detection_offline}: Mean average precision (\emph{mAP}), averaged across the test set classes, and Precision (\emph{P})/Recall (\emph{R}), with the operating point threshold determined by maximizing the \emph{F1} score on the validation set.

Each detection model uses a different category taxonomy: the ``Baseline'' and ``+SPI'' models output coarse-grained categories from the GPT, whereas ``+FPN'' and ``+MUL +CLN'' output fine-grained categories.
To ensure a fair comparison between all models, we roll up fine-grained categories to their corresponding coarse-grained categories in the offline detection evaluation.

With each detection iteration, we see steady gains in mAP, Precision, and Recall in Table~\ref{tab:detection_offline}.
``+SPI'' leads to a significant gain in mAP of +51\% relative on the ``Single-Product Images'' test set.
``+FPN'' improves mAP by +35\% relative on the ``Scene Images'' test set.
``+MUL +CLN'' results in a massive +102\% relative mAP gain on the ``Single-Product Images'' test set.
Many of these gains are a result of improved recall -- in particular with ``+FPN'', which agrees with the findings from \cite{DBLP:journals/corr/LinDGHHB16} that FPN significantly increases detection recall.

\begin{table}
\begin{center}
\begin{tabular}{l r r}
\hline
Model & Similar P@5 \\
\hline\hline
+SPI \footnotemark & 0.0\% \\
+FPN & +24.0\% \\
+MUL +CLN & +38.6\% \\
\hline
\end{tabular}
\end{center}
\caption{Detection model experiments on Home Decor E2E human relevance evaluations. We report relative gains: each gain is relative to the preceding row.}
\label{tab:detection_relevance}
\vspace{-5mm}
\end{table}
\footnotetext{Due to using Hybrid labelers (discussed further in Section~\ref{subsubsec:hybrid}) for end-to-end (E2E) relevance evaluation for the ``+SPI'' experiment, results were noisy, but fell in the same range as the previous model. Additionally, since E2E relevance evaluation is performed on scene query images, we do not expect ``+SPI'' to impact E2E Similar P@5. We conclude that ``+SPI'' had neutral impact on E2E Similar P@5. Note: Later detection experiments ``+FPN'' and ``+MUL +CLN'' used our in-house labeling team for reliable E2E relevance evaluations (discussed in Section~\ref{subsubsec:in_house}).}

\begin{table}
\begin{center}
\begin{tabular}{l r r}
\hline
Model & Scene Closeups & Product Click-throughs \\
\hline\hline
+SPI \footnotemark & - & - \\
+FPN & +51.6\% & +29.2\% \\
+MUL +CLN & +16.4\% & +3.2\% \\
\hline
\end{tabular}
\end{center}
\caption{Detection model experiment results from A/B experiment. We report relative gains: each gain is relative to the preceding row.}
\label{tab:detection_ab}
\vspace{-7mm}
\end{table}
\footnotetext{The ``+SPI'' model predated the launch of Shop The Look, so A/B experiment results were not measured.}




\subsubsection{\textbf{Human relevance evaluation}}
\label{subsubsec:detection_relevance}
We measure the impact of detection improvements on Shop The Look relevance using Home Decor E2E Similar P@5 in Table \ref{tab:detection_relevance}. The relevance gains are primarily due to improved predicted category label precision, since product results are filtered to the predicted category.

\subsubsection{\textbf{Online A/B experiment}}
The most significant user engagement gains are achieved with ``+FPN'': this model update leads to higher recall for objects in the home decor domain, translating to higher coverage. This increases both scene closeup volume and product click-through volume.
Results are shown in Table~\ref{tab:detection_ab}.

Similarly, ``+MUL +CLN'' leads to additional improvements in both scene closeup volume and product click-through volume.

%% file: main_embeddings.tex
\section{Candidate retrieval (embedding)}
\label{sec:embeddings}

After users tap on a query object from scene decomposition, Shop The Look recommends visually similar products from the Pinterest product corpus. We compute a visual embedding for the query object and use it to retrieve matching product candidates. The visual embedding we trained is an extension to our previous work on the multi-task high dimensional binary embedding \cite{unified-embedding}. 
In this section, we detail improvements in both our retrieval infrastructure, and our embedding model, to support the fashion shopping use case specifically.



 
\input{embedding_system.tex}

\subsection{Embedding improvements}
The baseline embedding \cite{unified-embedding} was only explicitly trained with one shopping use case -- home decor (\textit{Shop The Look Home}). In this section, we present how we improved visual embedding relevance for a new shopping use case -- fashion (\textit{Shop The Look Fashion}). We take full advantage of our multi-task embedding training framework. See Appendix~\ref{subsec:embedding_training_setup} for details on our embedding training setup.
\begin{enumerate}
    \item We observed general improvements in the literature from large-scale pretraining, so we update our backbone network with better pretrained weights (Section~\ref{subsubsec:pretrain}).
    \item To directly optimize for the fashion use case, we collect a separate set of fashion matching data and train our embedding model with new fashion matching tasks (Section~\ref{subsubsec:fashion_stl}).
    \item To address specific failure cases for fashion such as color and pattern mismatches, we include visual attribute prediction tasks in our embedding model (Section~\ref{subsubsec:attributes}).
\end{enumerate}

\subsubsection{\textbf{Large-scale pretraining}}
\label{subsubsec:pretrain}
We adopt large-scale pretrained wei\-ghts for our embedding model backbone network. Large-scale pretraining has been shown to help with various transfer learning tasks in the literature \cite{wsl-pretraining}, and it has also been proven in industry applications \cite{facebook-ad-wsl-pretraining}. Additionally, a higher capacity model is better at leveraging larger scale data \cite{wsl-pretraining}. Inspired by these findings, we update our backbone network from SE-ResNext101-32x4d \cite{senet} to ResNext101-32x8d \cite{resnext} pretrained with substantially larger data.

\subsubsection{\textbf{Fashion matching product dataset}}
\label{subsubsec:fashion_stl}
To expand Shop The Look to fashion, we collect a fashion matching product dataset to directly train our embedding for better fashion relevance. We start with 600,000 inspirational images from Pinterest creators, each annotated with a product they wish to promote with the inspirational image. Not all of the annotated products are of high quality -- wrong or missing bounding boxes, wrong categories, spam, and dissimilar product annotations are common problems. We clean up 600,000 image-product pairs with our label collection method detailed in Section \ref{sec:label_collection}. The resulting high-quality fashion matching product dataset has 113,000 images, with 9 coarse-grained category-level labels and 62,000 fine-grained instance-level labels. See Appendix~\ref{subsec:fashion_stl_cleanup} for further details on the cleanup process.

The collected fashion dataset is uniformly mixed into the training batch. Following the framework of multi-task embedding training, two additional cross-entropy classification losses are added to the embedding model for this dataset: one for the category label and one for the instance label. To mitigate the extreme classification problem, the instance label classification is subsampled to 2048 instances for each iteration. The method is detailed in \cite{unified-embedding}.


\input{embedding_attributes.tex}

\input{embedding_experiments.tex}

%% file: embedding_system.tex
\subsection{Retrieval infrastructure improvements}
Shop The Look is built on top of our in-house nearest neighbor retrieval system that also powers other visual discovery products at Pinterest \cite{pinterest-visual-search}. We present the modifications we made to our retrieval systems for Shop The Look:
\begin{enumerate}
    \item We started with a system that retrieves nearest neighbors purely based on visual embedding distance. To fix mistakes due to product results with mismatching category, we separate our retrieval indexes by category to support category restriction (Section~\ref{subsubsec:per_category_index}).
    \item We move to approximate nearest neighbor (ANN) search using locality-sensitive hashing (LSH) embedding tokens to leverage the full power of a search system (Section~\ref{subsubsec:ann_lsh}). This allows for restricting on additional attributes like gender, price, and domain, enabling further quality improvements and supporting product functionality. 
\end{enumerate}


\subsubsection{\textbf{Per-category Index}}
\label{subsubsec:per_category_index}

\begin{figure}[t]
\begin{center}
\includegraphics[width=0.7\linewidth]{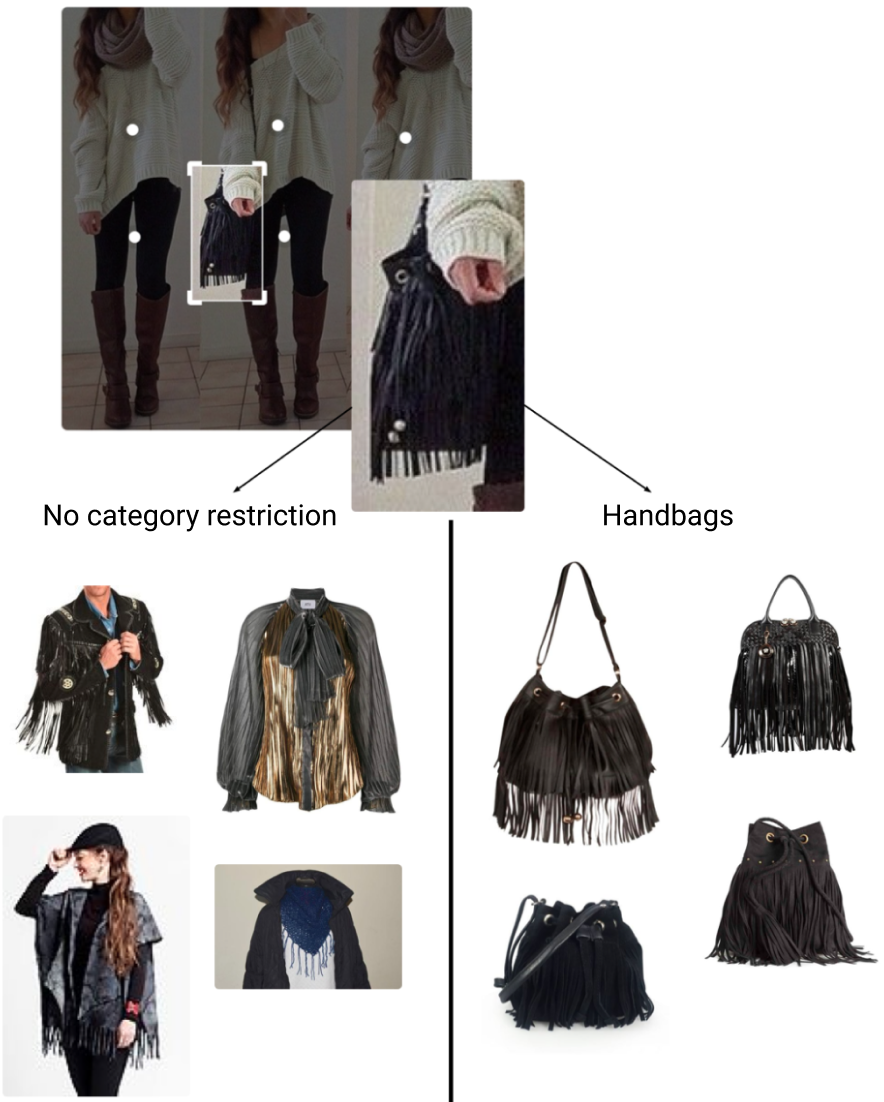}
\end{center}
\vspace{-3mm}
\caption{Visualization of retrieved results before (left) and after (right) applying category restrictions.}
\label{fig:category_restriction_vis}
\vspace{-5mm}
\end{figure}

Our first retrieval infrastructure was an approximate nearest neighbor search system which retrieves candidates from a tree index using Fast Library for Approximate Nearest Neighbors (FLANN) \cite{flann}. However, we discovered that retrieving candidates purely based on visual similarity may result in large semantic mismatch errors, such as returning candidates from a different category than the query object.

Simply using post-retrieval filtering by category cannot guarantee enough candidates with the desired category. Instead, to fix these mistakes, we add category as a restriction into retrieval. We first group candidates into different categories by using either an in-house Google Product Taxonomy (GPT) classifier or merchant-provided GPT labels. We build a separate index for each category. At retrieval time, we use the query object's predicted category label from scene decomposition to retrieve only candidates from the corresponding category index. 

This not only helps to eliminate many category mismatches, but also significantly reduces the search space of each query and therefore allows us to retrieve exact nearest neighbors instead of approximations. Figure~\ref{fig:category_restriction_vis} shows the qualitative difference in results before and after category restriction is applied. In an end-to-end (E2E) fashion relevance evaluation, after adding in category restriction, the number of "Bad" results in top-5 positions decreased by -17.1\%, while the number of "Similar" results in top-5 positions increased by +9.9\%.

\subsubsection{\textbf{Approximate Nearest Neighbor (ANN) using LSH embedding tokens}}
\label{subsubsec:ann_lsh}
As Shop The Look continued to evolve, there was an increasing need for more complex queries in addition to category. To support both improved quality and additional product functionality, we wanted to enable additional restrictions on other attributes like domain, gender, price, and merchant.

To fulfill these requirements, we adopt Pinterest's in-house ANN search platform \cite{manas} as the Shop The Look retrieval system. Each index includes an inverted index and a forward index. For each embedding, we use a LSH-based method to generate a set of tokens, which are used as keys to build the inverted index. The forward index stores the mapping from candidates' IDs to their embeddings.

For each query, we first retrieve candidates with matched tokens and rank them by the number of matches. The candidates are then reranked by calculating their embeddings' distance to the query embedding stored in the forward index.

In addition to embedding tokens, attribute values are also used as keys to build the inverted index. During retrieval, we send a tree-structured query to express complex search restrictions to the search engine, such as ``gender:Men \textbf{AND} (category:Shirt \textbf{OR} category:Tie) \textbf{AND} (\textbf{NOT} price < 50)''.

Gender is the first additional restriction we introduced into the retrieval system.
We index gender for all products in the Shop The Look corpus using an in-house gender classifier. Each product image is classified, or set to "unisex" if unsure. During retrieval, we predict the query image's gender and construct an ANN search query by using a combination of predicted query gender and "unisex" to prevent under-retrieval.
In an end-to-end (E2E) fashion relevance evaluation, after adding in gender restriction, the number of "Similar" results in top-5 positions increased by +5.9\%, while the number of "Extremely Similar" results in top-5 positions increased by +12.7\%. In an A/B experiment, product click-through rate decreased slightly by -0.49\%; however, we hypothesize this is likely due to noise, and judge it to be well outweighed by the relevance improvement.

%% file: embedding_attributes.tex
\subsubsection{\textbf{Visual fashion attributes}}
\label{subsubsec:attributes}

We observed common patterns of mistakes in fashion relevance evaluations, such as mismatching colors or patterns. One way to solve the issue is to collect more matching object-product pairs for each failure case, but it is difficult to come by such data. Label collection for classification is an easier task, so it can be accomplished more efficiently. Given this, we add fashion attribute classification data and tasks into embedding training to address these mistakes. To generate the attribute dataset, we first consult with fashion specialists to devise an attribute taxonomy and create labeler training guides; then collect labels on 70,000 image crops for the following attributes: Color, Pattern, Fabric, Lower body length, and Dress style; and finally, split the dataset into train / offline evaluation sets using a 90 : 10 split. We omit attribute labels that have too few samples. These attribute classification tasks are added to the embedding model architecture as additional tasks in our multi-task training system, each with one hidden fully-connected layer and softmax loss.

%% file: embedding_experiments.tex
\subsection{Embedding experiments}
\label{subsec:embedding_exps}
We show in the offline evaluations the complete evolution of our embeddings, each annotated with version number and its retrieval metric, and the results of selected versions in end-to-end human relevance evaluations and engagement A/B experiments.  For brevity, we report results on fashion relevance for embedding experiments because the improvements were focused on fashion.

\begin{figure}[t]
\begin{center}
\includegraphics[width=0.9\linewidth]{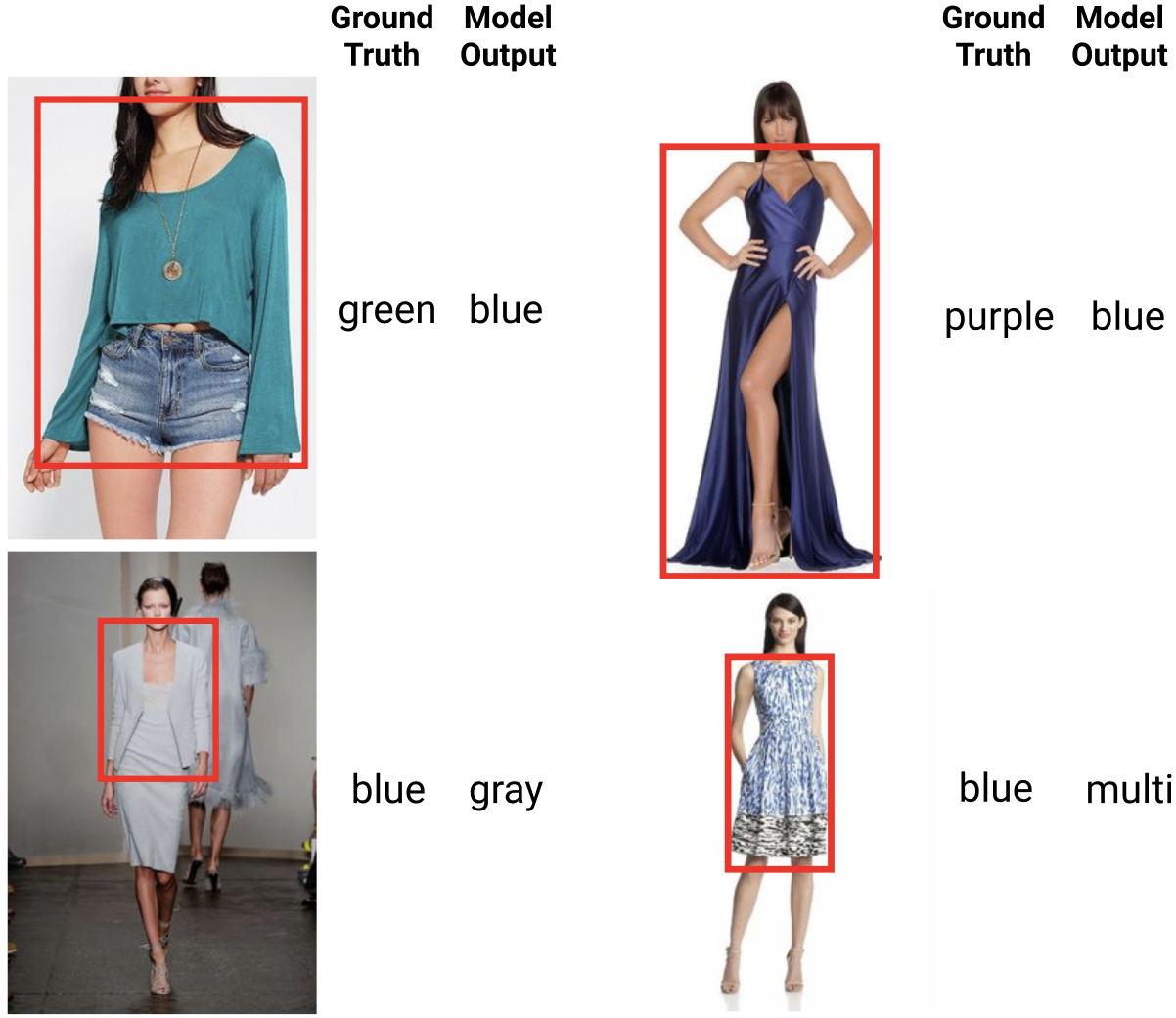}
\end{center}
\vspace{-3mm}
\caption{Visualization of mistakes made by Color attribute classifier in attribute offline evaluation. Many of these "mistakes" are actually ambiguous. Best viewed in color.}
\label{fig:attribute_mistakes_vis}
\vspace{-1mm}
\end{figure}

\subsubsection{\textbf{Offline evaluation}}
\label{subsubsec:embedding_offline_eval}
For offline retrieval evaluation, we collect a small matching product dataset for fashion (5,000 object-product pairs). In addition to the matching products, we also randomly sample 100,000 products from the Pinterest product corpus to form a distractor set. The ground truth products plus the distractors serve as our offline retrieval corpus. To simulate the online system, query objects are generated with the detector described in Section \ref{sec:decomposition}, and we compute binary embeddings for retrieval. Shop The Look Fashion Retrieval P@1 uses the standard top-1 precision in retrieval evaluation. Results are shown in Table ~\ref{tab:embedding_offline}.
We observe a significant improvement from ``+Fashion dataset'' (Section~\ref{subsubsec:fashion_stl}) of +79.8\% relative, and a smaller improvement from ``+Large pretrain'' (Section~\ref{subsubsec:pretrain}) of +6.5\% relative. 
To isolate the impact from attributes (Section~\ref{subsubsec:attributes}), we run an ablation experiment where we first remove all attributes from the ``V2'' embedding model (``\textit{-Attrs}''), and then add back only the Color classification task (``\textit{-Attrs} +Color''); this results in a +3.2\% relative improvement.

The attribute offline evaluation set follows the standard train/test split for classification dataset detailed in Section~\ref{subsubsec:attributes}. In Table~\ref{tab:attr_offline_eval}, we measure the accuracy of the top-1 predicted attribute (Attribute P@1) for each of the five attributes we collected. All attribute classifiers achieve >84\% precision for top-1 predicted label. Qualitatively, visualizations of attribute offline evaluations show that performance approaches human-level -- in many cases, flagged ``mistakes'' are in reality either mislabeled or due to ambiguity in the attribute taxonomy, stemming from unaddressed edge cases (Figure~\ref{fig:attribute_mistakes_vis}).

\begin{table}
\begin{center}
\begin{tabular}{l c c }
\hline
Model & Version & Shop The Look \\
 & & Fashion Retrieval P@1 \\
\hline\hline
Baseline & V1 & 0.323  \\
+Color & V1.1 & 0.325 \\
+Large pretrain & V1.2 & 0.346 \\
+Fashion dataset & V1.3 & 0.622 \\
+Pattern & V1.4 & 0.631 \\
+Fabric & V1.5 & 0.615 \\
+LL +DS & V2 & 0.641 \\
\hline
\textit{-Attrs} & & 0.620 \\
\textit{-Attrs} +Color & & 0.640 \\
\hline
\end{tabular}
\end{center}
\caption{Embedding retrieval experiments on offline evaluation, showing incremental additions. Reported metrics are measured with binarized embeddings. LL: Lower body length, DS: Dress style.}
\label{tab:embedding_offline}
\vspace{-8mm}
\end{table}

\begin{table}
\begin{center}
\begin{tabular}{l r r r r r}
\hline
Model & \multicolumn{5}{c}{Attribute P@1} \\
      & C & P & F & LL & DS \\
\hline\hline
V1.3 \textit{-Attrs} & - & - & - & - & - \\
+Color & 0.870 & - & - & - & - \\
+Pattern & 0.872 & 0.947 & - & - & - \\
+Fabric & 0.872 & 0.952 & 0.849 & - & - \\
+LL +DS & 0.872 & 0.949 & 0.848 & 0.898 & 0.879 \\
\hline
\end{tabular}
\end{center}
\caption{Attribute prediction offline evaluations from our embedding model. C: Color, P: Pattern, F: Fabric, LL: Lower body length, DS: Dress style.}
\label{tab:attr_offline_eval}
\vspace{-8mm}
\end{table}

\subsubsection{\textbf{Human relevance evaluation}}
\label{subsubsec:embedding_relevance}
We show embedding end-to-end (E2E) relevance results for fashion in Table~\ref{tab:embedding_relevance}. The gains are positively correlated with the offline evaluation metrics shown in Table ~\ref{tab:embedding_offline}. The largest improvement is from adding fashion training data as described in Section~\ref{subsubsec:fashion_stl} (See version ``V1.4''). We also see significant improvement from large-scale pretraining as described in Section~\ref{subsubsec:pretrain} and adding a fashion color attribute classification task as described in Section~\ref{subsubsec:attributes} (See version ``V1.2'').

\subsubsection{\textbf{Online A/B experiment}}
Since embedding improvements only affect product retrieval quality, we report product click-through rate in Table~\ref{tab:embedding_ab}. We observe that significant relevance gains from ``V1'' to ``V1.4'' translate to engagement gains.

\begin{table}
\begin{center}
\begin{tabular}{l r r}
\hline
Version & Similar P@5 & Extremely Similar P@5 \\
\hline\hline
V1.2 & +10.9\% & +38.1\% \\
V1.4 & +13.4\% & +54.7\% \\
V2 & -0.6\% & 0.0\% \\
\hline
\end{tabular}
\end{center}
\caption{Embedding model experiments on Fashion E2E relevance evaluations. We report relative gains: each gain is relative to the preceding row. The first row is relative to V1.}
\label{tab:embedding_relevance}
\vspace{-7mm}
\end{table}

\begin{table}
\begin{center}
\begin{tabular}{l r}
\hline
Version & Product Click-through Rate \\
\hline\hline
V1.2 & +7.14\% \\
V1.4 & +5.99\% \\
V2 & +0.83\% \\
\hline
\end{tabular}
\end{center}
\caption{Embedding model experiment results from A/B experiment. We report relative gains: each gain is relative to the preceding row. The first row is relative to V1.}
\label{tab:embedding_ab}
\vspace{-8mm}
\end{table}

%% file: main_label_collection.tex
\section{Label Collection}
\label{sec:label_collection}
Label collection is critical for both training/evaluation data collection and human relevance evaluation.

\emph{Training/Evaluation data collection.} There are three types of datasets we collect to improve system performance:

\begin{enumerate}
    \item For detection, we collect bounding boxes and category labels.
    \item For visual embeddings, we clean up creator-annotated fashion matching data by ensuring accuracy of bounding boxes, category labels, and matching products (Section~\ref{subsubsec:fashion_stl}).
    \item For attribute prediction, we collect attribute labels for each bounding box (Section~\ref{subsubsec:attributes}).
\end{enumerate}

\emph{Human relevance evaluation.} Human judges are used to evaluate Shop The Look relevance, as described in Section \ref{sec:evaluation}. To ensure the accuracy and consistency of the human labelers, we design home decor- or fashion-specific guidelines for assigning relevance ratings.






\input{label_collection_platform.tex}

%% file: label_collection_platform.tex
\subsection{Labeling platform improvements}
\label{subsec:labeler_platform}

Our goal is to get high-quality labels quickly and reliably. We observe that labeling platform heavily influences \textbf{label accuracy} (against a golden dataset containing expert labels) and \textbf{label consistency} (among labelers).
We investigate three labeling platforms:
\begin{enumerate}
    \item MTurk provides labels quickly and with low cost, but suffers from low consistency and accuracy (Section~\ref{subsubsec:mturk}).
    \item We attempt to improve on label consistency by proposing a fixed labeler pool with Hybrid (Section~\ref{subsubsec:hybrid}).
    \item To allow for real-time feedback from labelers and faster communication in pursuit of higher label accuracy and consistency, we propose an in-house labeling team with an on-site expert (Section~\ref{subsubsec:in_house}).
\end{enumerate}

\subsubsection{\textbf{Amazon Mechanical Turk (MTurk)}}
\label{subsubsec:mturk}
MTurk is optimized for speed and low cost. The labelers on MTurk do well on simple tasks such as bounding box labeling for limited categories. However, with more complicated tasks (e.g. E2E relevance evaluations), consistency and accuracy drop (Table~\ref{tab:label_platform_consistency}). The low consistency from labeler disagreement results in a low number of high-quality labels. Additionally, the labels we do get bias toward easier examples.

\subsubsection{\textbf{Third party fixed labeler pool (Hybrid)}}
\label{subsubsec:hybrid}
Given the low consistency and accuracy of MTurk, we propose using Hybrid, a platform with a fixed labeler pool. Label consistency improves slightly because we always use the same group of labelers (Table~\ref{tab:label_platform_consistency}). However, we observe a regression in label accuracy. The problem lies in the lack of feedback from labelers, making it difficult to disambiguate edge cases in order to refine the training guide.


\subsubsection{\textbf{In-house labeling team}}
\label{subsubsec:in_house}
To further increase label accuracy and consistency, we propose an in-house labeling team with a stable pool of labelers who can provide real-time feedback. In addition, we augment the team with an on-site expert who we train and calibrate with to refine the training guide. This on-site expert in turn trains, calibrates, and answers questions from labelers. The frequent and bilateral communication between the in-house labeling team and on-site expert, as well as between the on-site expert and our team, reduces the iteration time on the training guide, and thus speeds up the overall labeler training process. For high-stakes tasks, we can also check the accuracy of individual labelers and re-calibrate as necessary. The comprehensive training guide, together with a stable pool of labelers and an on-site expert, results in significant improvements in accuracy and consistency (Table~\ref{tab:label_platform_consistency}).

\begin{table}
\begin{center}
\begin{tabular}{l r r}
\hline
Labeling Platform & Label Consistency & Label Accuracy \\
\hline\hline
MTurk & 47\% & 32.5\% \\
Hybrid & 49\% & 26.0\% \\
In-house & 95\% & 89.5\% \\
\hline
\end{tabular}
\end{center}
\caption{Comparison of label consistency and accuracy across labeling platforms, for a sample home decor E2E relevance evaluation task. Label consistency is defined as the average of labeler agreement rates on the most common answers. Label accuracy is defined as percentage of labels that match the corresponding golden (expert) label.}
\label{tab:label_platform_consistency}
\vspace{-8mm}
\end{table}

%% file: main_lessons.tex
\section{Lessons Learned}
\label{sec:lessons}

\subsection{Label subjectivity is challenging}
Many tasks, such as fashion or home decor similarity, are seemingly subjective; yet, subjectivity in labels is detrimental to relevance evaluation and training data collection, as it leads to poor label quality. How then can we ensure objective labels for those tasks? We found that \textit{label inconsistency} was a symptom of labeling subjectivity, and \textit{edge cases} were a common cause of labeling subjectivity. In the following, we present examples of each, along with our mitigations.

\subsubsection{\textbf{Label inconsistency}}
We tackled label inconsistency by moving to a dedicated in-house labeler team. It helped greatly by (1) enabling improved consistency amongst different labelers through calibration sessions, where an on-site expert guides the labeler team through difficult examples, and (2) increasing labeler accountability, since labels are now traceable back to individuals for further calibration.


\subsubsection{\textbf{Edge cases}}
From visualizing labeling results, we discovered that inconsistent labels were often due to edge cases that hadn't been addressed in labeler training guides. Our mitigation strategies generally involved discovering ambiguous examples, and disambiguating similar cases in the labeler training guide. For instance, moving to an in-house labeling team as our labeling platform allowed for direct feedback from labelers on hard examples.
Additionally, we found that "mistakes" in offline evaluations were also useful in finding instructive edge cases, e.g. in the attribute offline evaluation (Figure~\ref{fig:attribute_mistakes_vis}). This leads to a virtuous cycle of iterating on labeler training guides, collecting cleaner training data to further refine the model, and using more accurate offline evaluations to discover even harder edge cases.

\begin{figure}[t]
\begin{center}
\includegraphics[width=1.0\linewidth]{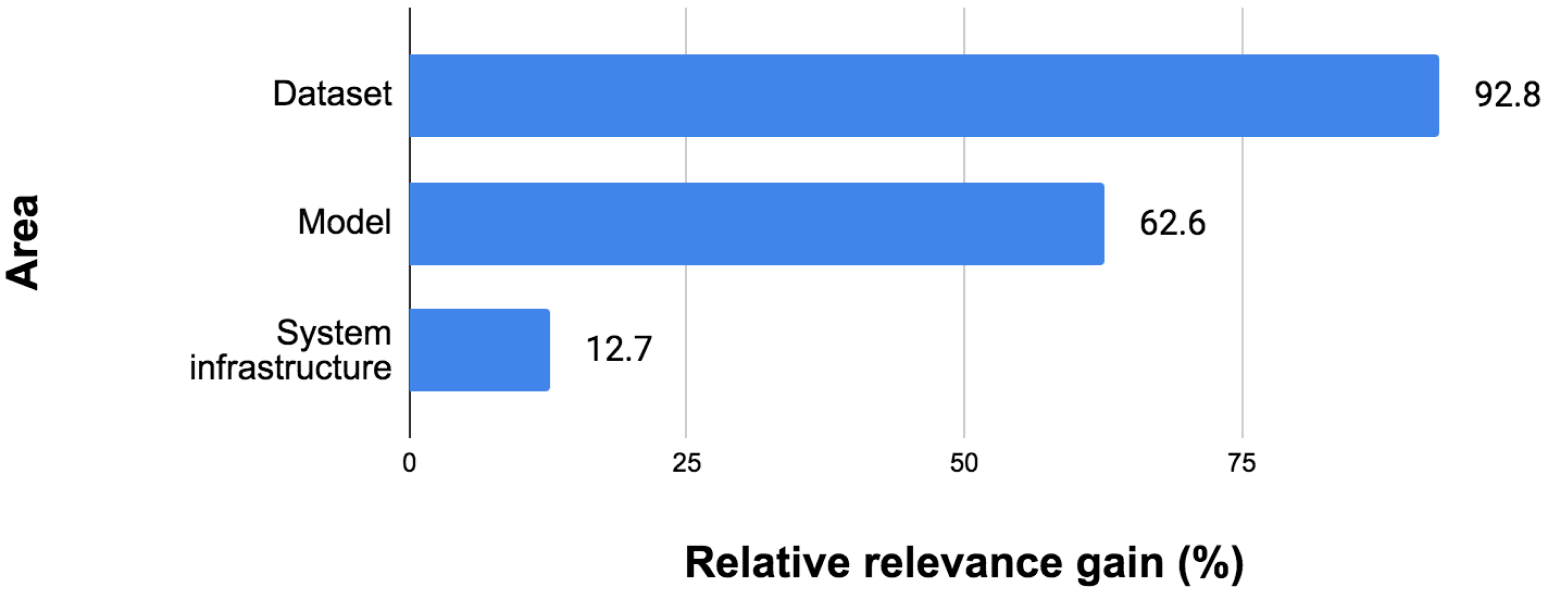}
\end{center}
\vspace{-3mm}
\caption{Breakdown of cumulative relative relevance gains by area.}
\label{fig:impact_breakdown}
\vspace{-3mm}
\end{figure}


\subsection{Impact comes from diverse areas}
In Shop The Look, we observed significant cumulative gains of over 160\% in E2E human relevance judgements and over 80\% in engagement from all of the following areas: \textit{dataset}, \textit{model}, and \textit{system infrastructure}. A relative impact breakdown for E2E relevance evaluations is shown in Figure~\ref{fig:impact_breakdown}; we calculated this based on Home Decor Similar@5 data for detection experiments from Table~\ref{tab:detection_relevance} and Fashion Extremely Similar P@5 data for embedding experiments from Table~\ref{tab:embedding_relevance} (ideally we would target an Extremely Similar metric in Home Decor as well, but the Home Decor relevance evaluation that was available when running detection experiments only had a coarser measure of Similar / Not Similar). In this breakdown, dataset improvements had the largest impact, followed by model and then system infrastructure. Therefore, in addition to improving the model and infrastructure, we also invest heavily in the label collection platform.


\subsubsection{\textbf{Dataset improvements}}
These tended to be straightforward, but often led to significant wins. For detection, adding the single-product image dataset into training significantly improved mAP in the single-product offline evaluation. For embeddings, adding fashion matching product data (ensuring that fashion training data exists for the desired fashion task), large-scale pretraining, and Color and Pattern attributes led to large gains in fashion E2E relevance and product click-through engagement rate.

\subsubsection{\textbf{Model improvements}}
We found that upgrading to the latest detection architecture led to large wins in scene mAP, relevance, and engagement (both scene closeup and product click-through volume). Enabling multi-scale image augmentation resulted in significant gains in both scene and single-product mAP, as well as relevance and scene closeup engagement volume.

\subsubsection{\textbf{System infrastructure improvements}}
Infrastructure advancements can unlock gains when it might be difficult to do so via modeling work. For instance, we observed that embedding-based retrieval often struggles with out-of-category and out-of-gender results, so moving to per-category retrieval indexes and then our in-house ANN search infrastructure enabled us to implement these as restrictions during retrieval, removing the burden of encoding category and gender information into the embedding. Restricting search by category greatly decreased large semantic mismatch errors and increased fashion relevance, while restricting based on gender led to significant improvements in fashion relevance.

%% file: main_conclusion.tex
\vspace{-1mm}

\section{Conclusion}

Shop The Look is a visual shopping system at Pinterest, which allows our users to discover and purchase products within images. Shop The Look performs scene decomposition using an object detector and candidate retrieval using visual embeddings, providing users a seamless shopping experience from inspirational images to products they can shop. With improvements in scene decomposition, our object detector discovers more shoppable objects in a greater number of inspirational images, which increases the chances for users to get Shop The Look recommendations. With the developments in candidate retrieval, we improve the quality of products recommended to users, either by reducing bad results such as category mismatches, or by improving the visual similarity of product results. The improvements we discuss would not be possible without our substantial investment in label collection. With our in-house labeling platform, we reduce subjectivity in label collection by focusing on inconsistency among labelers and disambiguating edge cases. It allows us to bilaterally communicate our requirements for fast, consistent, and objective label collection. In this work, we provide a holistic view of a complex system to maximize impact, based on observing significant improvements from diverse areas such as training data, model, and system infrastructure. We hope our learnings will not only continue to improve the shopping experience for our users, but also provide valuable insights for other practitioners in the field.


%% file: main_acknowledgements.tex
\vspace{-1mm}

\section{Acknowledgements}

We thank Mariellen Barros for driving labeling efforts; Joyce Zha, Amanda Strickler, and AJ \"Oxendine for frontend support; Jen Chan for product management support during the initial release of Shop The Look; Yiming Jen for helping with the migration to our in-house ANN search infrastructure; Josh Beal for providing a signal used to generate the single-product dataset; Nick DeChant for supporting embedding and detection object extraction infrastructure; Michael Mi and Zheng Liu for building and supporting our in-house ANN search infrastructure; Kofi Boakye, Dmitry Kislyuk, Bin Shen, and the three anonymous reviewers for their insightful comments, which helped to greatly clarify and improve this paper.

%% file: appendix.tex
\section{Appendix on Reproducibility}

\subsection{Detection training setup}
\label{subsec:dtn_setup}
We utilize the Caffe \cite{jia2014caffe} framework to train the ResNet101-Fast\-erRCNN baseline models, and the Detectron \cite{Detectron2018} Caffe2-based framework to train the ResNeXt101-FasterRCNN-FPN models. We train our object detection models on a compute machine with eight Tesla V-100 GPUs.
For models trained in Caffe, we use a single GPU with a batch size of 1, which is the default setting.
For models trained in Detectron, we utilize a batch size of two images per graphics card, for a total effective batch size of 16, and use synchronized SGD.

\subsection{Single-product dataset generation}
\label{subsec:single_product_generation}
We detail the process of generating the single-product dataset in Algorithm~\ref{alg:single_product_generation}: ``pinterest\_product\_corpus'' is a list of products in the Pinterest corpus, each with an image, category, and flag for whether the category is provided by the merchant or predicted using an in-house classifier; ``scene\_distribution'' maps from category to number of examples in the scene detection dataset; ``single\_product\_dataset'' is a list of tuples (image, category, largest bounding box) representing examples in the single-product dataset.

\subsection{Embedding training setup}
\label{subsec:embedding_training_setup}
The embedding model is trained using PyTorch on one p3dn.24xlarge Amazon EC2 instance with eight Tesla V100 graphics cards. We use all 8 GPUs on the machine for DistributedDataParallel training, and we also use Nvidia's Apex library\footnote{https://github.com/NVIDIA/apex} for mixed precision training.

The hyperparameters we use are mostly the same as~\cite{unified-embedding}. We replace our base model with ResNeXt101\_32x8d~\cite{resnext}. We use SGD with momentum of 0.9, weight decay of 1e-4, and gamma of 0.1. We start the training with base model frozen, and train the newly initialized parameters for 2 epochs with learning rate 0.6. We then train end-to-end with a batch size of 128 per GPU and apply gamma to reduce learning rate every 5 epochs for a total of 15 epochs of end-to-end training. During training, we apply horizontal mirroring, random crops, and color jitter to resized 256x256 images, while during testing we center crop to a 224x224 image from the resized image.

Compared to the multi-task setting described in ~\cite{unified-embedding}, our work has two additional datasets: fashion and attributes. We find that because the attribute dataset has smaller size, and the tasks are easier, choosing a task weight hyperparameter of 0.1 for all the attribute classification tasks achieves better overall offline evaluation performance.

\begin{algorithm}
\caption{Single-product dataset generation}\label{alg:single_product_generation}
\begin{flushleft}
\textbf{Input:} pinterest\_product\_corpus, scene\_distribution\\
\textbf{Output:} single\_product\_dataset\\
\end{flushleft}
\begin{algorithmic}[1]
\STATE{$raw\_dataset \leftarrow list([])$}
\FORALL{$product \in pinterest\_product\_corpus$}
    \IF{$has\_merchant\_provided\_category(product)$}
        \STATE{$img \leftarrow get\_image(product)$}
        \STATE{$cat \leftarrow get\_category(product)$}
        \IF{$is\_product\_on\_white\_background(img)$}
            \STATE{$bounding\_boxes \leftarrow run\_detection(img)$}
            \STATE{$largest\_bbox \leftarrow max\_by\_area(bounding\_boxes)$}
            \IF{($largest\_bbox.width \times largest\_bbox.height) \geq 0.8 \times (img.width \times img.height)$}
                \STATE{$raw\_dataset.add((img, cat, largest\_bbox))$}
            \ENDIF
        \ENDIF
    \ENDIF
\ENDFOR

\STATE{$single\_prod\_distribution \leftarrow map(\{\})$}
\STATE{$single\_product\_dataset \leftarrow list([])$}
\FORALL{$(img, cat, largest\_bbox) \in raw\_dataset$}
    \IF{$scene\_distribution.contains(cat)$}
        \STATE{$limit \leftarrow scene\_distribution.get(cat)$}
        \STATE{$num \leftarrow single\_prod\_distribution.get(cat, default=0)$}
        \IF{$num < limit$}
            \STATE{$single\_prod\_distribution.set(cat, num+1)$}
            \STATE{$single\_product\_dataset.add((img, cat, largest\_bbox))$}
        \ENDIF
    \ENDIF
\ENDFOR

\RETURN{$single\_product\_dataset$}
\end{algorithmic}
\end{algorithm}

\subsection{Fashion matching product dataset cleanup}
\label{subsec:fashion_stl_cleanup}
At Pinterest, we have a collection of inspirational images annotated with products, which Pinterest creators wish to promote. We first use our in-house classifier (based on textual, visual and engagement information) to filter the inspirational images by men's and women's fashion. The resulting 600K image-product pairs are then sent for manual cleanup.

The cleanup process is done by our in-house labeling team, as discussed in Section~\ref{subsubsec:in_house}. Image-product pairs are presented to the labelers. If the product is in the image, the labeler annotates the coarse fashion category (out of 9 possible categories), and draws bounding boxes around all the occurrences of the product in the image. The cleaned up dataset has 97,000 high quality matching pairs, which constitute our fashion matching product dataset.